\documentclass[letterpaper, 10 pt, conference]{ieeeconf}  
\IEEEoverridecommandlockouts
\usepackage{cite}
\usepackage{amsmath,amssymb,amsfonts}
\usepackage{algorithmic}
\usepackage{graphicx}
\usepackage{textcomp}
\usepackage{booktabs}
\usepackage{xcolor}
\usepackage{longtable}
\usepackage{tabularx}
\usepackage{url}
\bstctlcite{MyBSTcontrol}
\title{Hand-centric Human-to-Robot Trajectory Transfer from Video Demonstrations via Object Category-agnostic Temporal Localization 
\thanks{Karlsruhe Institute of Technology, Karlsruhe, Germany. Email: {\tt\small \{name\}.\{surname\}@kit.edu}.}
\thanks{This work is supported by the German Federal Ministry of Research, Technology, and Space (BMFTR) under the Robotics Institute Germany (RIG), the DFG SFB-1574-471687386 project, and the Ministry of Science, Research and Arts of the Federal State of Baden-Württemberg within the InnovationCampus Future Mobility}
}

\author{
Yitian Shi$^*$, Di Wen$^*$, Zhengqi Han, Zicheng Guo, Yu Hu, Edgar Welte, \\ Kunyu Peng, Rainer Stiefelhagen, Rania Rayyes 
}

\begin{document}
\maketitle

\bstctlcite{MyBSTcontrol}
\begin{abstract}
   Human videos provide a scalable source of demonstrations for robot imitation learning. 
   However, converting them into executable and semantically aligned robot trajectories requires determining \emph{when} task-relevant interactions occur, \emph{how} demonstrated human grasps should be retargeted across embodiments, and \emph{which} post-contact motions should be reproduced.
    In this paper, we present \textit{HOCALT} (\emph{H}and--\emph{O}bject \emph{C}ategory--\emph{A}gnostic \emph{L}ocalization and \emph{T}ransfer), a hand-centric retargeting framework that unifies 3D hand motion reconstruction, temporal contact localization, and cross-embodiment trajectory generation.
    Given synchronized stereo videos, \textit{HOCALT} identifies task-relevant contact intervals by jointly reasoning over reconstructed 3D hand articulation and category-agnostic object motion cues.
    These intervals serve as temporal anchors for initializing cross-embodiment transfer, where the demonstrated human grasps are retargeted into multi-modal robot grasp hypotheses.
    Within each interval, these hypotheses are then propagated along the demonstrated hand motion, yielding executable trajectories.
    Finally, we augment the transferred trajectories to generate diverse variants from a single demonstration.
    Across various tasks, \emph{HOCALT} outperforms VLM-based temporal localization baselines, achieving higher replay success than existing retargeting approaches. 
\end{abstract}

\section{Introduction}

In imitation learning, distilling demonstrations from human videos may significantly reduce the human effort for manual teleoperation and kinesthetic teaching~\cite{YOTO, R+X, ZeroMimic}. 
Although human video demonstrations provide a scalable source of manipulation data, preserving contact-rich Hand--Object Interaction (HOI) structure remains challenging due to noisy hand motion estimates, uncertain contact timing, and embodiment mismatch. 
For parallel-jaw (PJ) end-effectors, existing human-to-robot transfer methods commonly rely on simplified representations such as sparse keypoint-based templates~\cite{RTAGrasp, R+X, zhang2026robowheel} or object-centric affordances~\cite{ZeroMimic}, which may amplify hand pose errors, discard demonstration-specific grasps, or limit generalization to unseen objects.

\begin{figure}[h!]
\centering    
\includegraphics[width=.98\linewidth]{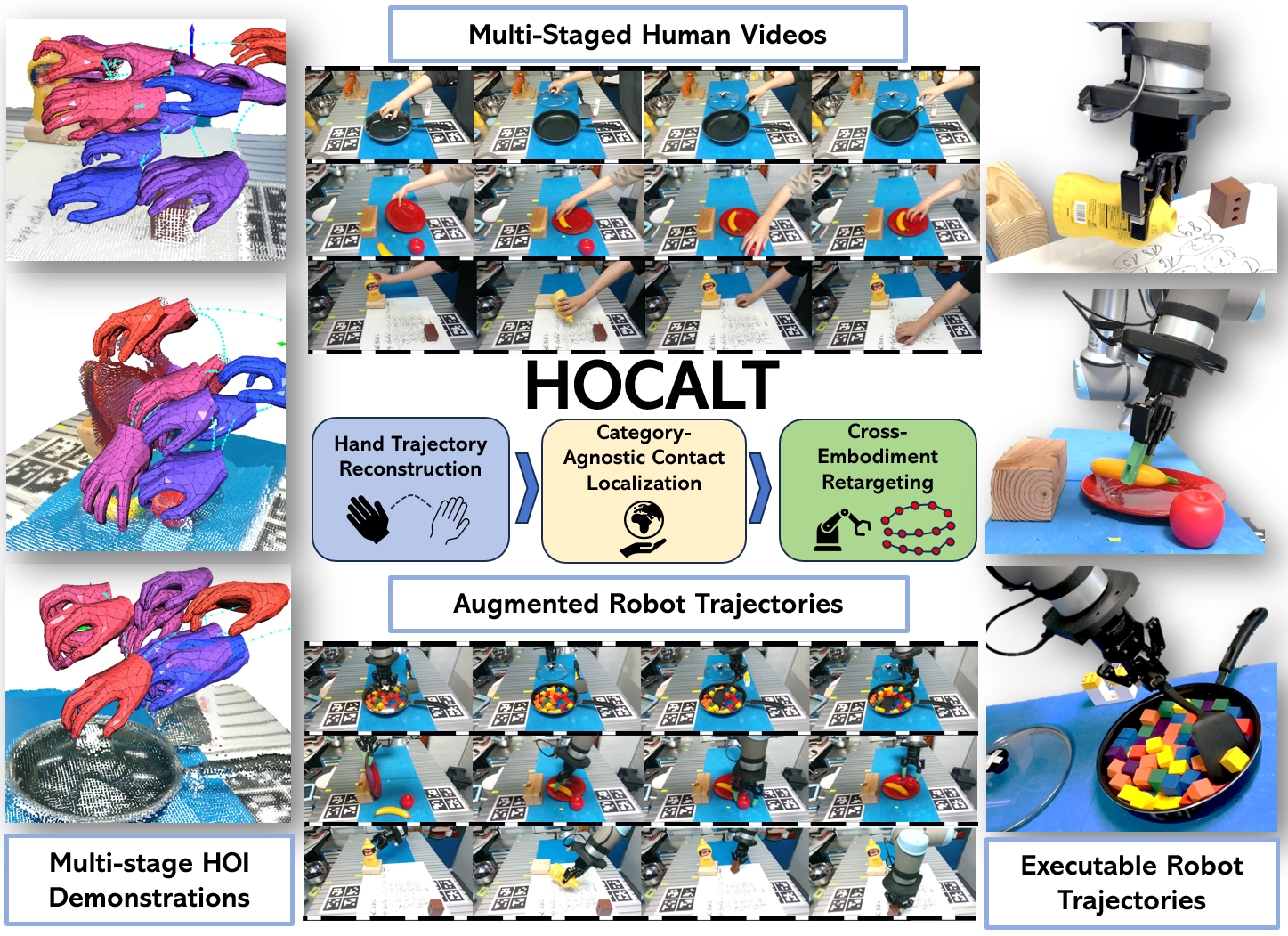}
\caption{From a single multi-view human demonstration, HOCALT localizes task-relevant interactions, retargets human grasps, and transfers the corresponding hand motion into diverse executable PJ trajectories.}
    \label{fig:placeholder}
\end{figure}

Moreover, spatial retargeting and temporal action localization are tightly coupled throughout the interaction transfer. This leads to a further challenge in determining \emph{when} task-relevant interactions occur and \emph{which} temporal segments should be transferred.  
Human videos often include long approach motions, idle pauses, and repeated transitions between contact and release, making it challenging to isolate the task-relevant manipulation intervals. 
Existing approaches~\cite{wake2025open, EgoLoc} rely on inference-time prompting of Vision-Language Models (VLMs) to infer interaction boundaries, while precise localization remains challenging in repetitive or long-horizon demonstrations containing multiple interaction stages.

To address these challenges, we present \emph{HOCALT}, a hand-centric retargeting framework that converts stereo human video demonstrations into diverse and interaction-aware robot trajectories. 
Starting from the reconstructed metric 3D wrist motion and MANO~\cite{MANO} hand articulation, \emph{HOCALT} employs a novel two-stage, object category-agnostic temporal localizer that identifies task-relevant contact intervals by jointly reasoning over hand motion and manipulated object cues. 
Then, each localized interval serves as a basic unit of transfer: at its onset, the pretrained \emph{HOGraspFlow}~\cite{HOGraspFlow} retargets the observed human hand configuration into multi-modal PJ grasp hypotheses. 
Each hypothesis is propagated by the demonstrated wrist motion into an in-contact end-effector trajectory within the interval. 
In this way, our temporal localizer provides both the initial grasp anchors and relevant motion segments required for trajectory transfer, without instance-level labels or pose tracking on the manipulated object. Finally, \emph{HOCALT} augments the interval-level trajectories and connects consecutive intervals through collision-free motion planning to produce complete executable episodes.

In summary, our contributions are threefold:
(i) \textbf{Hand-centric trajectory transfer.} We propose a hand-centric framework that transfers contact timing, grasp structure, and task-relevant motion into executable PJ trajectories without semantic labels, object meshes, or pose tracking.
(ii) \textbf{Category-agnostic temporal contact localization.} We introduce a novel temporal contact localizer that combines hand articulation with interval-specific object motion cues, which identify task-relevant contact intervals without online VLM prompting;
(iii) \textbf{Interval-specific trajectory augmentation.} We propose a waypoint-based trajectory augmentation strategy that diversifies retargeted trajectories while preserving local motion structure and executability.

\section{Related Work}

\subsection{Transferring Manipulation from Human Videos}
Human videos are retargeted either to supervise policy learning~\cite{MotionTracks,PointPolicy,Track2Act,ZeroMimic,YOTO} or to condition robot execution at deployment~\cite{R+X,park2025demodiffusion}.
In the former setting, Motion Tracks~\cite{MotionTracks}, Point Policy~\cite{PointPolicy}, and Track2Act~\cite{Track2Act} bridge the embodiment gap through transferable 2D motion tracks, 3D keypoints, and predicted point trajectories, respectively. 
ZeroMimic~\cite{ZeroMimic} learns skill-specific policies from task-conditioned grasp affordances and post-grasp wrist motion, while YOTO~\cite{YOTO} converts human motion into keyframe-based robot trajectories and expands them through automated replay.
A distinct line of work uses human demonstrations as in-context conditioning during policy runtime.
R+X~\cite{R+X} conditions action generation on hand motion exemplars, which are retrieved by task-specific VLM prompts, while DemoDiffusion~\cite{park2025demodiffusion} refines retargeted human hand trajectories using a pretrained robot policy.
These methods primarily rely on predefined geometric abstractions on embodiments, which limits their robustness to complex, noisy videos involving diverse hand motions and multi-stage interactions.

In contrast, \emph{HOCALT} localizes hand--object contact intervals to anchor multi-modal grasp retargeting and trajectory augmentation. Hence, it generates diverse executable trajectories without relying on a fixed cross-embodiment mapping, thereby improving the utility of a single video demonstration.


\subsection{Temporal Localization and Grasp Retargeting}

Transferring a human demonstration requires determining \emph{when} hand--object contact occurs and \emph{how} the demonstrated grasp maps to the target embodiment. 
Existing approaches typically address these aspects separately: For temporal localization, T-PIVOT~\cite{wake2025open} localizes open-vocabulary action segments through iterative VLM prompting, whereas EgoLoc~\cite{EgoLoc} targets contact and separation timestamps using hand motion-guided sampling and VLM refinement;
For embodiment retargeting, hand-centric grasp transfer infers semantic-aware grasp affordances from human demonstrations~\cite{RTAGrasp, GAT-Grasp}, while relying on pre-defined matching between human and robot end-effectors; 
Morphology-based retargeting enables contact-consistent and kinematic-aware grasps across embodiments~\cite{GeometryMatching, khargonkar2025robotfingerprint}, but depends on explicit object models or simulation-based optimization~\cite{Spider}.

Instead, \emph{HOCALT} uses contact intervals to couple hand-centric grasp retargeting with in-contact motion transfer, while relying on Vision Foundation Models (VFMs)~\cite{lin2025depth, SAM3} for geometric and motion-based interaction reasoning with local deployment rather than online VLM prompting.

\section{Methodology}
\begin{figure*}
    \centering
    \includegraphics[width=.96\linewidth]{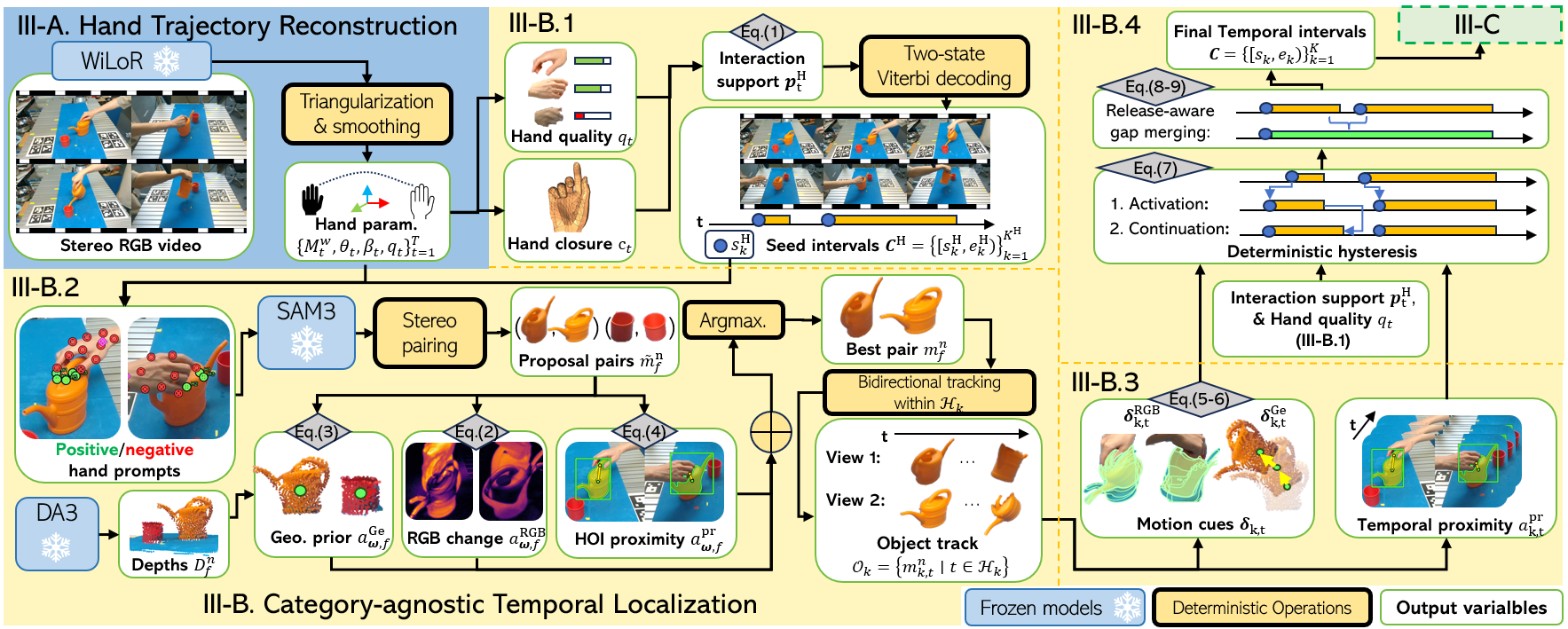}
    \caption{Architecture of \textit{HOCALT} including Hand Trajectory Reconstruction (Sec.~\ref{sec:hand_pose_estimation}) and Category-agnostic Temporal Localization (Sec.~\ref{sec:instance_agnostic_contact})}
    \label{fig:pipeline}
\end{figure*}
Given a synchronized stereo RGB video with human demonstration, \textit{HOCALT} converts the demonstrated hand--object interactions into executable PJ trajectories in three stages: First, it reconstructs and smooths metric 3D hand motion (Sec.~\ref{sec:hand_pose_estimation}); 
Second, within initial seed intervals derived from hand motion, our category-agnostic temporal localizer discovers and tracks the actively manipulated object. The geometric and object cues from these tracks are then used to refine the contact boundaries (Sec.~\ref{sec:instance_agnostic_contact}); Third, at each localized contact onset, we use pretrained \emph{HOGraspFlow} \cite{HOGraspFlow} to generate multi-modal PJ grasp hypotheses. HOCALT subsequently propagates these hypotheses using the reconstructed wrist motion and augments them into executable trajectory variants (Sec.~\ref{sec:trajectory_retargeting}). 

\subsection{Hand Trajectory Reconstruction}
\label{sec:hand_pose_estimation}
As illustrated in Fig.~\ref{fig:pipeline} (blue), given a synchronized stereo RGB sequence $\mathcal{V}=\{(I_t^1,I_t^2)\}_{t=1}^{T_0}$, where $T_0$ is the input length and $I_t^n$ is frame $t$ from view $n\in\{1,2\}$,
we use WiLoR~\cite{WILOR}, a vision-based foundational hand reconstructor to estimate the wrist orientation $R_t^n\in SO(3)$, articulated joint pose $\theta_t^n\in\mathbb{R}^{45}$, and MANO shape parameters $\beta_t^n\in\mathbb{R}^{10}$ in the $t$-th frame. 
We consider the frame $t$ as valid only when the WiLoR confidence scores of hand detections $q_t^n\in[0,1]$ on both views exceed a fixed threshold.


For each valid stereo frame, we triangulate the corresponding MANO joints to recover the metric wrist pose $\tilde M_t^w\in \mathrm{SE}(3)$ in the world frame. The per-view estimates are fused through geodesic/arithmetic averaging of rotations and the remaining parameters, respectively. 
For short gaps between valid observations, we linearly interpolate the MANO parameters and subsequently smooth the wrist trajectory using an IEKF–RTS smoother~\cite{IEKF-RTS}. After trimming the unobserved leading and trailing frames, we re-index the remaining sequence, yielding the reconstructed hand sequence
$\{M_t^w,\theta_t,\beta_t,q_t\}_{t=1}^{T}$.

\subsection{Category-agnostic Temporal Contact Localization}
\label{sec:instance_agnostic_contact}
Given the reconstructed hand motion $\{M_t^w,\theta_t,\beta_t,q_t\}_{t=1}^{T}$, our temporal contact localizer aims to estimate a set of contact intervals $\boldsymbol C=\{[s_k,e_k)\}_{k=1}^{K}$, where $s_k$ and $e_k$ denote the contact and release frames of the $k$-th interaction. 
We assume one active hand and at most one actively manipulated object within each contact interval, and different intervals may select different objects without strict identity continuity.


Overall, the localizer first derives hand-driven seed intervals of potential contact (Sec.~\ref{sec:hand_driven_first_pass}). For each seed interval, it generates candidate proposals, selects the one most likely to represent the actively manipulated object, and tracks the corresponding object (Sec.~\ref{sec:interaction_object_discovery}). It then extracts motion cues from the resulting object track (Sec.~\ref{sec:object_motion_cue}) and combines them with hand cues to refine the interval boundaries (Sec.~\ref{sec:second_pass_phase_decoding}). The entire process is illustrated in Fig.~\ref{fig:pipeline} (yellow).


\subsubsection{Hand-driven First Pass}
\label{sec:hand_driven_first_pass}
Before the manipulated object is identified, our first pass aims to obtain coarse contact intervals from hand articulation and their visibility across views. 
Given the fused articulation vector \(\theta_t\) and confidence score \(q_t\) from Sec.~\ref{sec:hand_pose_estimation}, we define the hand closure score as \(c_t=\lVert\theta_t\rVert_2/\sqrt{45}\), motivated by the observation that finger flexion commonly accompanies contact onset.

To obtain robust interaction support from noisy $c_t$, let $\mathcal R_t(\boldsymbol y)$ denote the normalized value of a continuous scalar signal $\boldsymbol y$ at frame $t$. This normalizer maps its temporal 10th and 90th percentiles to $[0, 1]$, and clips values outside this range. We calculate the interaction support $p_t^{\mathrm H}$ by
\begin{equation}
\kappa_t=\mathcal R_t(\boldsymbol c),
\quad
p_t^{\mathrm H}=
\mathcal S(\kappa_t \cdot q_t).
\label{eq:hand_support}
\end{equation}
with $\mathcal S$ as the second-order Savitzky--Golay smoother \cite{savitzky1964smoothing} with frame window length as 7. Here, $q_t$ downweights the hand support with poor hand confidence.

To suppress isolated score fluctuations from the frame-wise support score $\{p_t^{\mathrm H}\}_{t=1}^{T}$, we leverage a two-state Viterbi decoder to determine whether the interaction is active at each frame. Consecutive active frames are then grouped into $K^{\mathrm H}$ seed intervals: $\boldsymbol C^{\mathrm H}=\{C_k^{\mathrm H}\}_{k=1}^{K^{\mathrm H}}$. These intervals are used for category-agnostic object discovery in Sec.~\ref{sec:interaction_object_discovery}.

\subsubsection{Interaction-grounded Object Discovery}
\label{sec:interaction_object_discovery}
For each seed interval $C_k^{\mathrm H}=[s_k^{\mathrm H},e_k^{\mathrm H})$, SAM3 generates candidate object masks in a small set of frames neighbouring the onset frame $s_k^{\mathrm H}$.
Specifically, points near the fingertips are used as positive prompts, while projected MANO joints away from the contact region are used as negative prompts to suppress hand masks.
Candidate masks from SAM3 are paired across views using stereo matching and cross-view consistency.

For the $\omega$-th candidate pair $(\tilde m_{\omega, f}^1,\tilde m_{\omega, f}^2)$ at proposal frame $f$, we compute three complementary cues that measure its likelihood of being the actively manipulated object.

First, to analyze the appearance-based object cue from video, we first measure the \emph{photometric changes} within the proposal masks. Let $\rho_{\omega, f}^n\in[0, 1]$ denote the average ratio of RGB change within the mask $\tilde m_{\omega, f}^n$ over nearby frames. Since genuine object motion should be captured by both viewpoints, we penalize the cross-view inconsistency caused by potential occlusion or segmentation errors from a single viewpoint,  yielding the photometric object cue $a_{\omega, f}^{\mathrm{RGB}}$:
\begin{equation}
a_{\omega, f}^{\mathrm{RGB}}=\frac{\rho_{\omega, f}^1+\rho_{\omega, f}^2}{2}-\frac{|\rho_{\omega, f}^1-\rho_{\omega, f}^2|}{4}.
\label{eq:rgb_change_proposal}
\end{equation}

Since photometric changes may respond to background artifacts, we additionally assess the cross-view geometric consistency of each candidate pair. A valid stereo pair should back-project to a compact and coherent 3D cluster in the world frame. We therefore define the \emph{geometric prior} $a_{\omega,f}^{\mathrm{Ge}}$:
\begin{equation}
\begin{aligned}
\mathcal X_{\omega, f}^n
&=\Pi_n^{-1}(\tilde m_{\omega, f}^n,D_f^n), \quad \boldsymbol\mu_{\omega, f}=\operatorname{median}_{\mathbf x\in\mathcal X_{\omega, f}^1 \cup\mathcal X_{\omega, f}^2}\mathbf x\\
a_{\omega, f}^{\mathrm{Ge}}
&=\exp\!\left[-\frac{
\mathrm P_{90}(\{\lVert\mathbf x-
\boldsymbol\mu_{\omega, f}\rVert_2\}_{\mathbf x\in\mathcal X_{\omega, f}^1 \cup\mathcal X_{\omega, f}^2 })}
{0.06}\right].
\end{aligned}
\label{eq:da3_proposal_prior}
\end{equation}
Specifically, $D_f^n$ is the depth map estimated by Depth Anything 3~\cite{lin2025depth}, whose masked area $\tilde m^n_{\omega,f}$ is back-projected into the world coordinate via $\Pi_n^{-1}$ to get the 3D point set $\mathcal X_{\omega, f}^n$. $P_{90}$ denotes the $90$th percentile. Hence, $a_{\omega, f}^{\mathrm{Ge}}$ assigns higher scores to stereo pairs that form a compact 3D cluster.

As the last cue, we measure $a_{\omega, f}^{\mathrm{pr}}$ as the \emph{hand--object proximity} in the image space. Let $\mathbf h_{\omega, f}^n$ denote the median coordinate of the projected MANO joints,
$\mathbf z_{\omega, f}^n$ the centroid of the candidate mask $\tilde m_{\omega, f}^n$,
and $\ell_{\omega, f}^n$ the diagonal length of its bounding box.
We calculate
\begin{equation}
a_{\omega, f}^{n,\mathrm{pr}}
=
\exp\!\left(
-\frac{\|\mathbf h_{\omega, f}^n-\mathbf z_{\omega, f}^n\|_2}{\ell_{\omega, f}^n}
\right), \, a_{\omega, f}^{\mathrm{pr}} = \frac{a_{\omega, f}^{1,\mathrm{pr}}+a_{\omega, f}^{2,\mathrm{pr}}}{2}
\label{eq:hand_object_proximity}
\end{equation}

Among all candidates across the proposal frames, we sum the \emph{photometric change} $a_{\omega, f}^{\mathrm{RGB}}$, \emph{geometric prior} $a_{\omega, f}^{\mathrm{Ge}}$, and \emph{hand-object proximity} $a_{\omega, f}^{\mathrm{pr}}$ and select the pair with the highest total score. We then use SAM3 to track the selected object bidirectionally over the tracking horizon \(\mathcal H_k\), which extends around \(C_k^{\mathrm H}\) with valid stereo hand detections. This produces the interval-specific object track \(\mathcal O_k=\{m_{k,t}^n\mid t\in\mathcal H_k,\ n\in\{1,2\}\}\), from which we derive the motion cue to refine the boundaries of seed intervals in Sec.~\ref{sec:object_motion_cue}.

\begin{figure*}
    \centering
\includegraphics[width=\linewidth]{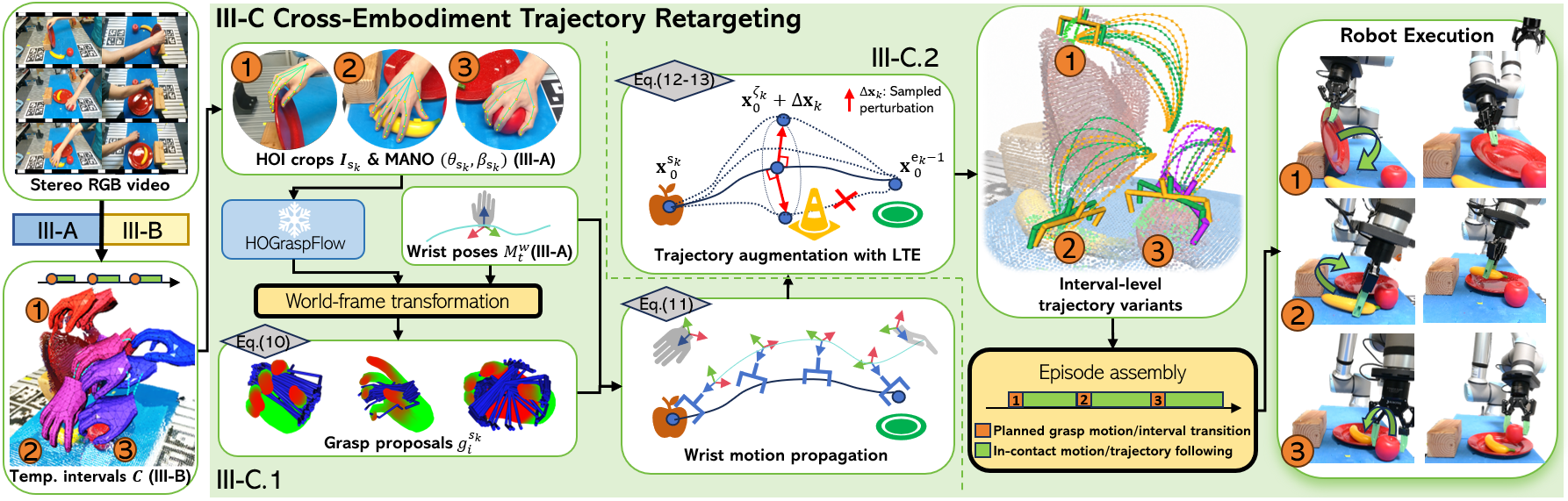}
    \caption{Cross-embodiment transfer from \textit{HOCALT} (I) queries the pretrained \emph{HOGraspFlow}~\cite{HOGraspFlow} for taxonomy-aware grasp hypotheses, (II) propagates and augments their trajectories, (III) assembles multi-stage robot episodes, and (IV) robot execution for evaluation or data collection.}
    \label{fig:cross-embodiment-trajectory-transfer}
\end{figure*}

\subsubsection{Object Motion Cue}
\label{sec:object_motion_cue}
Given the interval-specific object track $\mathcal O_k$, we analyze its temporal motion to provide complementary evidence of continued manipulation to the seed interval. Specifically, for each selected object track \(\mathcal O_k\), similar to \ref{sec:interaction_object_discovery}, we quantify its motion cue over $\mathcal H_k$ from both photometric and geometric perspectives.

Let $\{f_j\}_{j=1}^{J}\subseteq\mathcal H_k$ denote the $J$ sampled frames.
At each frame $f_j$, we reuse the back-projection in Eq.~\eqref{eq:da3_proposal_prior} to obtain the median of 3D point sets $\mathbf \mu_{k,f_j}$  averaged over viewpoints.
The geometric motion cue is calculated by
\begin{equation}
\delta_{k,f_j}^{\mathrm{Ge}}
=1-\exp\!\left(
-\frac{\lVert \mathbf \mu_{k,f_j}-\mathbf \mu_{k,f_{j-1}}\rVert_2}
{\sigma_{\text{3D}}\cdot(f_j-f_{j-1})}
\right)
\end{equation}
$\sigma_{\mathrm{3D}}=0.006$ is a fixed motion scale.
The sparsely sampled scores \(\delta_{k,f_j}^{\mathrm{Ge}}\) are interpolated over \(\mathcal H_k\) and smoothed to obtain the frame-wise geometric motion score \(\delta_{k,t}^{\mathrm{Ge}}\).

Similarly, we measure the photometric motion score $\delta_{k,t}^{\mathrm{RGB}}$ via Eq.~\eqref{eq:rgb_change_proposal}.
We then define the unified object motion support:
\begin{equation}
\delta_{k,t}=\mathcal S\!\left(
\max\left(
\delta_{k,t}^{\mathrm{RGB}},
s_d\cdot
\delta_{k,t}^{\mathrm{Ge}}
\right)\right).
\label{eq:object_motion_support}
\end{equation}
Here, $s_d>0$ controls the contribution of the DA3 cue. A high \(\delta_{k,t}\) indicates potential continued object motion, based on either photometric changes or estimated 3D displacement.

\subsubsection{Second-pass Phase Decoding}
\label{sec:second_pass_phase_decoding}

For each seed interval $C_k^{\mathrm H}$, the second pass combines
the hand support $p_t^{\mathrm H}$ from    Eq.~\eqref{eq:hand_support} with the object
motion trace $\delta_{k,t}$ from
Eq.~\eqref{eq:object_motion_support} to refine its boundaries.
In addition, we apply Eq.~\eqref{eq:hand_object_proximity} to each tracked mask in $\mathcal{O}_k$ and average the per-view scores to obtain the proximity cue $a_{k,t}^{\mathrm{pr}}$.
Our hysteresis decoder then applies the following rules for boundary refinement sequentially:
\begin{equation}
\begin{aligned}
\text{Activation:}&\
\max(
p_t^{\mathrm H},
q_t\cdot a^{\text{pr}}_{k,t}
)\geq\tau_c, \ t\in[1, \ldots, T], \\
\text{Continuation:}&\
\max(
p_t^{\mathrm H},
q_t\cdot a^{\text{pr}}_{k,t}
)\geq\tau_r
\,\lor\,\delta_{k,t}\geq\tau_\delta,
\end{aligned}
\label{eq:second_pass_decoding}
\end{equation}
where $\tau_c>\tau_r$, and both $a_{k,t}^{\mathrm{pr}}$ and $\delta_{k,t}$ set to
$0$ outside $\mathcal H_k$.

Concretely, the decoder first applies the activation rule over the complete sequence to construct an initial active mask. Then, for each anchored seed interval $C_k^{\mathrm H}$, it extends the onset $s_k^{\mathrm H}$ backward till $\hat s_k$ if the preceding mask value stays active. At the same time, the decoder extends the endpoint $e_k^{\mathrm H}$ forward to $\hat e_k$ when the continuation rule holds.

 
After obtaining the provisional intervals $\hat {C_k}=[\hat s_k,\hat e_k)$, to prevent brief interruptions in the continuation cue from spuriously splitting a single interaction, we further examine the gaps between successive provisional intervals. 

For $\hat C_k=[\hat s_k,\hat e_k)$, let
$
G_k=[\hat e_k,\hat s_{k+1})
$ be the nonempty intermediate gap between
$\hat C_k$ and $\hat C_{k+1}$.
Since object motion may persist after release, we complement the continuation rule with a separate release cue:
\begin{equation}
r_t=
\mathbf 1\!\left[
p_t^{\mathrm H}<0.25
\land
v_t\geq0.25
\right],
\end{equation}
where $v_t \in [0, 1]$ is the normalized and smoothed frame-wise wrist displacement. The merge decision is
\begin{equation}
M(G_k)=
\mathbf 1\!\left[
|G_k|\leq g_{\max}
\land
\frac{1}{|G_k|}
\sum_{t\in G_k} r_t
\leq 0.5
\right],
\label{eq:interval_merge}
\end{equation}
where $g_{\max}=10$.
Thus, a gap is merged when it contains insufficient cues of an actual release. If $M(G_k)=1$, the two provisional intervals are merged into $[\hat s_k,\hat e_{k+1})$. After merging, the remaining intervals are re-indexed as the final temporal intervals
$\boldsymbol C=\{[s_k,e_k)\}_{k=1}^{K}$.

\subsection{Cross-Embodiment Trajectory Retargeting}
\label{sec:trajectory_retargeting}

As illustrated in Fig.~\ref{fig:cross-embodiment-trajectory-transfer}, given the smoothed wrist trajectory \(\{M_t^w\}_{t=1}^{T}\) from Sec.~\ref{sec:hand_pose_estimation} and localized contact intervals \(\boldsymbol C\) from Sec.~\ref{sec:second_pass_phase_decoding}, we retarget the human grasp at each contact onset into multiple PJ grasp hypotheses and propagate them along the wrist motion within the interval (Sec.~\ref{sec:grasp_retarget}). The resulting interval-level trajectories are subsequently augmented and assembled through collision-free motion planning to form complete robot episodes (Sec.~\ref{sec:traj_augment}).


\subsubsection{Grasp retargeting}
\label{sec:grasp_retarget}
For a contact interval $C_k$, we employ \emph{HOGraspFlow}~\cite{HOGraspFlow} to retarget the demonstrated human grasp intent into  PJ grasp hypotheses. Specifically, given the hand-centered RGB crop $ I_{s_k}^{\mathrm H}$ obtained from the WiLoR detection and the hand parameters $\{\theta_{s_k},
\beta_{s_k}\}$, \emph{HOGraspFlow} generates grasp hypotheses from a conditional distribution over $\mathrm{SE}(3)$:
\begin{equation}
\tilde g_{k,i}\sim
p_{\phi}\left(
g\mid
 I_{s_k}^{\mathrm H},
\theta_{s_k},
\beta_{s_k}
\right),
\quad
g_{k,i}^{s_k}=M_{s_k}^w \tilde g_{k,i}.
\end{equation}
Since \emph{HOGraspFlow} produces multi-modal grasp hypotheses, we cluster the sampled $50$ grasps using DBSCAN~\cite{ester1996density} with the geodesic distance metric on \(\mathrm{SE}(3)\) to remove isolated predictions and retain dominant ones. 

For a sampled grasp $g_{k,i}^{s_k}$ and wrist pose $M_t^w$, the wrist propagation throughout frames returns
\begin{equation}
\Delta M_{k,t}
=
M_t^w \left(M_{s_k}^w\right)^{-1},\quad g_{k,i}^{t}= \Delta M_{k,t}g_{k,i}^{s_k}.
\end{equation}
This propagation preserves the wrist-relative grasp transform and reproduces the demonstrated in-contact hand motion in the robot end-effector space.

\subsubsection{Trajectory Augmentation and Episode Assembly}
\label{sec:traj_augment}
To generate diverse trajectories from a single demonstration, we perturb an intermediate control point approximately orthogonal to the local motion direction and use Laplacian Trajectory Editing (LTE)~\cite{nierhoff2016spatial} to smoothly propagate the perturbation while preserving local motion structure. This process is illustrated in Fig.~\ref{fig:cross-embodiment-trajectory-transfer}

Let
$\mathbf X_0=[\mathbf x_0^{s_k},\ldots,\mathbf x_0^{e_k-1}]^\top$
denote the propagated translation sequence. We select an intermediate control point at \(\zeta_k\in[s_k,e_k)\), with the perturbation magnitude scaled by the endpoint displacement:
\begin{equation}
\ell_k
=
||
\mathbf x_0^{e_k-1}-\mathbf x_0^{s_k}
||_2,
\quad
\rho_k
\sim
\mathcal U(0.15\ell_k,0.25\ell_k)
\end{equation}
We then sample a perturbation $\Delta\mathbf x_k$ orthogonal to the local trajectory direction at $\zeta_k$, with $\left\lVert\Delta\mathbf x_k\right\rVert_2=\rho_k$. With the discrete Laplacian operator denoted as $L$, we aim to solve:
\begin{equation}
\begin{aligned}
&\hat{\mathbf X}
=
\arg\min_{\mathbf X}\;
||
L\mathbf X-L\mathbf X_0
||_F^2
+
\lambda_s
||
\mathbf x^{s_k}-\mathbf x_0^{s_k}
||_2^2
\\
+ &
\lambda_\zeta
||
\mathbf x^{\zeta_k}
-
(
\mathbf x_0^{\zeta_k}+\Delta\mathbf x_k
)
||_2^2
+
\lambda_e
||
\mathbf x^{e_k-1}-\mathbf x_0^{e_k-1}
||_2^2 .
\end{aligned}
\end{equation}
Here, \(\lambda_s\), \(\lambda_\zeta\), and \(\lambda_e\) weight the start point, perturbed control point, and endpoint constraints, respectively. 
%
The solution \(\hat{\mathbf X}\) defines the augmented translation sequence, which is combined with the original propagated orientations to form an augmented trajectory.

Finally, to assemble complete episodes, collision-free motion planning connects the terminal pose at \(e_k-1\) of each interaction to the next grasp pose at \(s_{k+1}\). These planned free-space motions are interleaved with the augmented trajectories constructed from \(\hat{\mathbf X}\), forming complete multi-stage robot episodes for replay, evaluation, or data collection.


\section{Experiments}

Our evaluation addresses 3 central questions: (i) Can our temporal contact localizer capture task-relevant contact intervals accurately across single- and multi-stage demonstrations? (ii) Can \textit{HOCALT} generate physically executable trajectories, and how do they compare to teleoperation? (iii) Can the transferred trajectories provide effective supervision across different imitation learning algorithms?

\begin{figure}
    \centering
    \includegraphics[width=.95\linewidth]{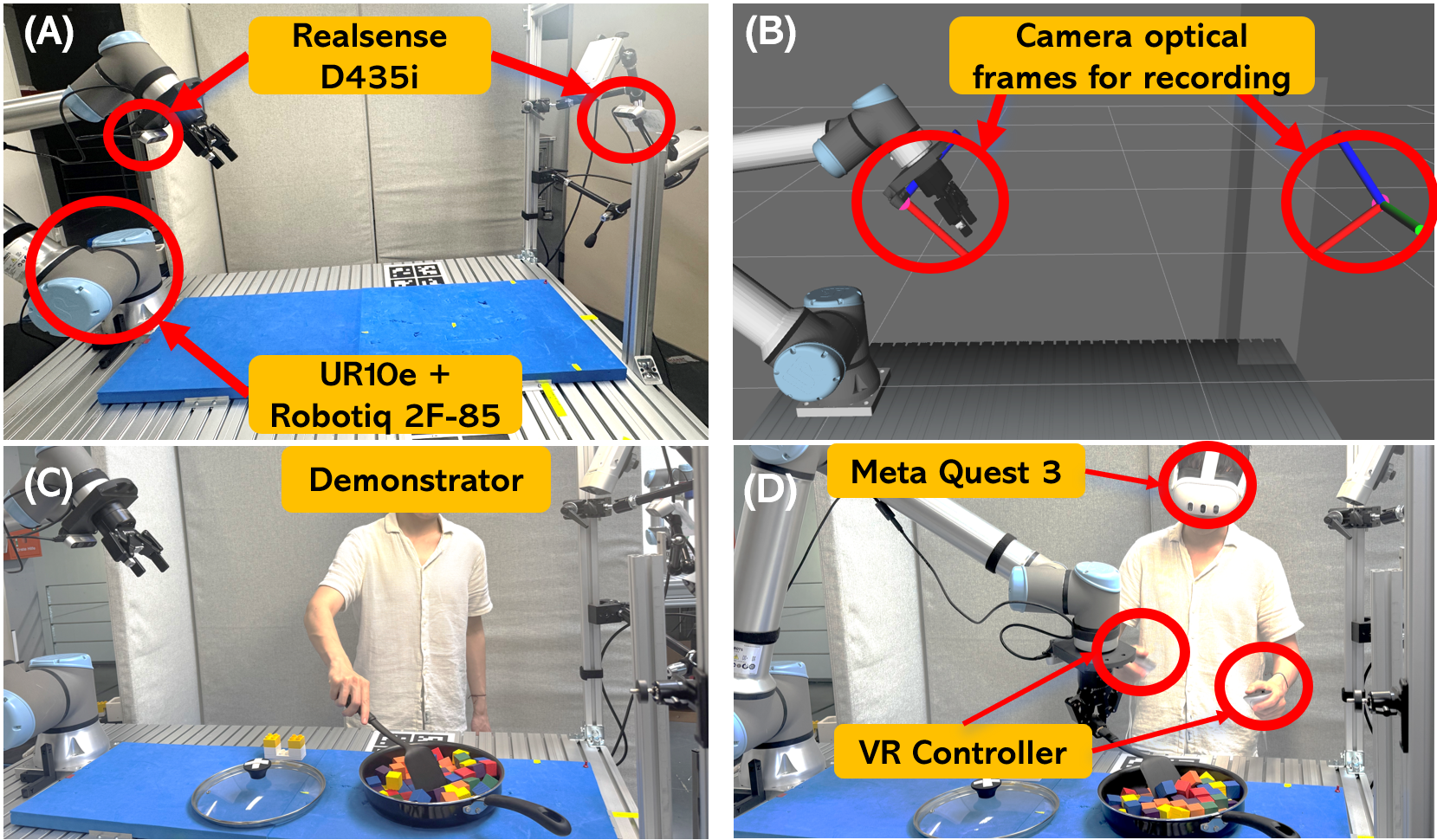}
    \caption{Experimental setup. (A) Shared workspace for human demonstrations, replay, and policy inference, equipped with a UR10e robot, a Robotiq 2F-85 gripper, and 2 Intel RealSense D435i cameras. (B) Calibrated optical frames of the two cameras. (C) Human demonstration recording. (D) Teleoperation setup using a Meta Quest 3 headset and VR controllers.}
    \label{fig:hardware}
\end{figure}

\begin{figure*}
    \centering
    \includegraphics[width=\linewidth]{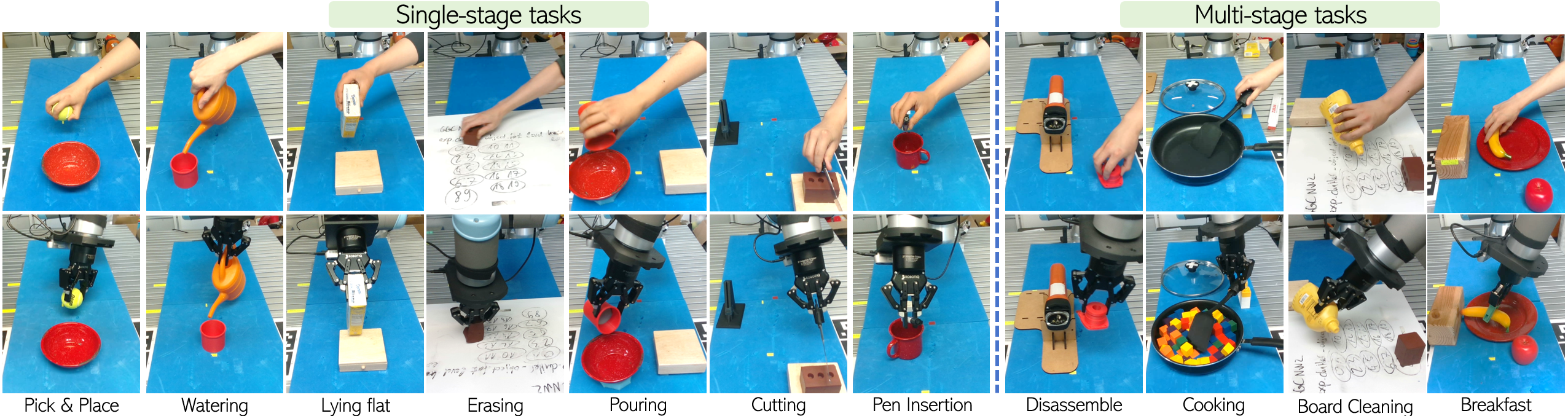}
    \caption{Task set in multi-object workspaces, including 7 single-stage tasks and 4 multi-stage tasks. In Multi-stage tasks, (i) \emph{Disassemble} needs to remove the red part and put down the angle grinder; (ii) \emph{Cooking} needs to stir the cubes after the lid is taken away; (iii) \emph{Board Cleaning} needs to pour the detergent before erasing; and (iv) \emph{Breakfast} needs to put down the plate before placing the fruits in order.}
    \label{fig:tasks}
\end{figure*}

To answer these questions, we construct a benchmark comprising 110 synchronized stereo demonstrations across 11 tasks, with 10 demonstrations per task (illustrated in Fig.~\ref{fig:tasks}). All recordings contain multiple objects in the workspace, and each is recorded at 30 fps, ranging between 20 and 40 seconds. Among the tasks, 7 are single-stage, while 4 contain ordered multi-stage interactions that may manipulate different task-relevant objects across intervals. For evaluation of the temporal contact localizer, ground-truth contact intervals were manually annotated and cross-checked by 2 independent annotators to reduce boundary ambiguity.
Fig.~\ref {fig:hardware} shows the hardware setups for data collection.

\subsection{Temporal Contact Localization}

\noindent\textbf{Baselines of overall comparison}
We compare against 3 baselines: \textit{Threshold} is a hand-geometry heuristic. 
In each view, it labels a frame as contact when the reconstructed 3D distance between the thumb and index fingertips falls below 6~cm. As VLM-based baselines, \textit{T-PIVOT}~\cite{wake2025open} iteratively refines action boundaries selected by VLM from uniformly sampled frames; 
\textit{EgoLoc}~\cite{EgoLoc} instead uses weighted sampling of candidate frames that prefers low 3D hand velocities. 
We use GPT-4o~\cite{hurst2024gpt} as the underlying VLM for both methods. 
Although \emph{EgoLoc} was originally developed for egocentric videos, it has also been evaluated on an exocentric benchmark. Its wrist-velocity analysis requires 3D hand poses expressed in a common reference frame, a requirement naturally satisfied by our fixed-camera setup.
Here we use \emph{EgoLoc} with Self-Adaptive Sampling Strategy (SASS) for long videos as the specific baseline.  
For a fair comparison with these single-view approaches, we also report \emph{Ours (Mono)}, which processes the same monocular demonstrations. Compared to \emph{Ours}, no stereo wrist trajectory or cross-view matching is used.

\noindent\textbf{Evaluation metrics} Following EgoLoc~\cite{EgoLoc}, we evaluate contact onset and release localization using success rate (SR) and mean absolute error (MAE). Within each video, predicted and annotated intervals are matched one-to-one in descending order of interval IoU. SR($\omega$), where $\omega\in\{3,5,10\}$ frames, measures the fraction of all annotated onset and release boundaries whose matched predictions have an absolute error of at most $\omega$ frames, and unmatched annotated boundaries are counted as failures. MAE is computed over matched onset and release boundaries only. In addition, we compare frame-wise IoU, precision, recall, and F1 score.

\noindent\textbf{Overall results} Table~\ref{tab:contact_localization_overall} shows that \textit{Ours} achieves the best overall temporal contact localization across metrics.
Notably, \textit{Ours (Mono)} already achieves an MAE of 10.62 frames and an F1 score of 0.89, substantially outperforming all baselines. 
This advantage arises from a persistent object-conditioned cue rather than hand posture or action semantics alone. 
\textit{Threshold}'s high recall but low precision indicates that finger closure also occurs during approach and pauses. 
The VLM baselines identify plausible action frames but do not maintain object identity: \textit{T-PIVOT} may miss the target event during coarse sampling, while \textit{EgoLoc}'s low velocity prior also responds to intermediate pauses, such as those in pouring and cutting. 

\begin{table}[t]
\centering
\setlength{\tabcolsep}{2pt}
\caption{Overall temporal localization results}
\label{tab:contact_localization_overall}
\begin{tabular}{l|cccccccc}
\toprule
Approach & SR(3)$\uparrow$ & SR(5)$\uparrow$ & SR(10)$\uparrow$ & MAE$\downarrow$ & Rec.$\uparrow$ & IoU$\uparrow$ & Prec.$\uparrow$ & F1$\uparrow$ \\
\midrule
\textit{Threshold} & 0.36 & 0.42& 0.51& 30.20& 0.78& 0.47& 0.51& 0.58 \\
\textit{T-PIVOT\cite{wake2025open}}
& 0.07 & 0.12 & 0.24 & 42.03
& 0.64 & 0.48 & 0.66 & 0.60 \\
\textit{EgoLoc\cite{EgoLoc}} & 0.16 & 0.26 & 0.37 & 34.61 & 0.66 & 0.51 & 0.71 & 0.65 \\
\midrule
\textit{Ours (Mono)} & 0.50 & 0.57 & 0.71 & 10.62 & 0.88 & 0.82 & \textbf{0.93} & 0.89 \\
\textit{Ours} &
\textbf{0.54} &
\textbf{0.65} &
\textbf{0.77} &
\textbf{8.34} &
\textbf{0.93} &
\textbf{0.85} &
0.91 &
\textbf{0.92} \\
\bottomrule
\end{tabular}
\end{table}

\begin{table}[t]
\centering
\setlength{\tabcolsep}{3.pt}
\caption{Ablation study on the temporal contact localizer modules}
\label{tab:temporal_ablation}
\begin{tabular}{l|cccccc}
\toprule
Variant & SR(10)$\uparrow$ & MAE$\downarrow$ & Rec.$\uparrow$ & IoU$\uparrow$ & Prec.$\uparrow$ & F1$\uparrow$ \\
\midrule
\multicolumn{7}{c}{\textit{A: Interaction-grounded object discovery (Sec. \ref{sec:interaction_object_discovery})}} \\
\midrule
First pass only & 0.68 & 11.99 & 0.79 & 0.76 & 0.96 & 0.85 \\
w/o \emph{geometric prior} & 0.73 & 9.42 & 0.93 & 0.85 & 0.91 & 0.91 \\
w/o \emph{photometric change} & 0.74 & 9.92 & 0.93 & 0.84 & 0.91 & 0.91 \\
w/o \emph{hand-object proximity} & 0.73 & 10.90 & 0.92 & 0.86 & 0.92 & 0.91 \\
\midrule
\multicolumn{7}{c}{\textit{B: Second-pass phase decoding (Sec. \ref{sec:second_pass_phase_decoding})}} \\
\midrule
w/o RGB motion & 0.72 & 10.55 & 0.89 & 0.81 & 0.91 & 0.89 \\
w/o DA3 motion & 0.73 & 10.43 & 0.88 & 0.82 & 0.92 & 0.89 \\
w/o release cue in merge & 0.71 & 11.89 & 0.92 & 0.83 & 0.90 & 0.90 \\
\midrule
\textit{Ours} & 0.77 & 8.34 & 0.93 & 0.85 & 0.91 & 0.92 \\
\bottomrule
\end{tabular}
\end{table}

\begin{figure}
    \centering
    \includegraphics[width=\linewidth]{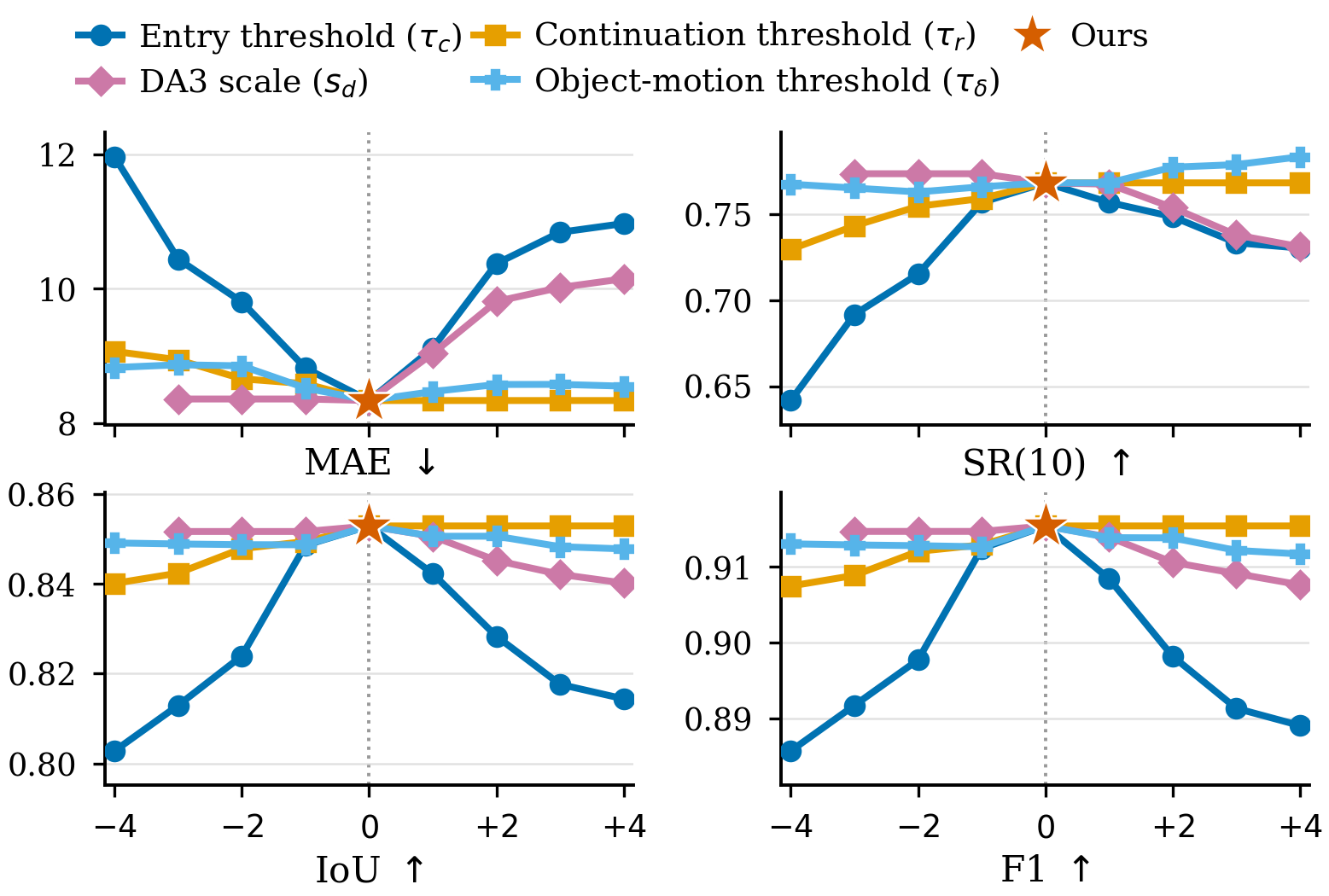}
    \caption{Sensitivity test on parameters from Eq.\eqref{eq:object_motion_support} and Eq.\eqref{eq:second_pass_decoding}. Parameter intervals are reported in Table~\ref{tab:sensitivity_intervals}. The x-axes are in step units. }
    \label{fig:temporal_sensitivity}
\end{figure}

\begin{table}[t]
\centering
\caption{Parameter intervals for the analysis in Fig.~\ref{fig:temporal_sensitivity}.}
\label{tab:sensitivity_intervals}
\setlength{\tabcolsep}{2pt}
\begin{tabular}{cccc|cccc}
\toprule
Param. & Baseline & Range & Step &
Param. & Baseline & Range & Step \\
\midrule
$\tau_c$
& 0.55 & $[0.35,\,0.75]$ & 0.05 &
$\tau_r$
& 0.35 & $[0.15,\,0.55]$ & 0.05 \\
$s_d$
& 0.75 & $[0.00,\,1.25]$ & 0.25 &
$\tau_\delta$
& 0.55 & $[0.35,\,0.75]$ & 0.05 \\
\bottomrule
\end{tabular}%
\end{table}

\noindent\textbf{Ablation studies}
To investigate where the gains in temporal contact localization originate, we conduct controlled ablations on the two main stages of our localizer. 
Specifically, Group A ablates interaction-grounded object discovery: (i) \emph{First pass only} directly uses the hand-derived seed intervals \(\boldsymbol C^{\mathrm H}\) without object discovery; (ii-iv) Three variants removes the \emph{geometric prior} \(a_{\omega,f}^{\mathrm{Ge}}\), \emph{photometric change} \(a_{\omega,f}^{\mathrm{RGB}}\) and \emph{hand-object proximity} \(a_{\omega,f}^{\mathrm{pr}}\)  from proposal ranking individually; Moreover, group B fixes the resulting object tracks \(\mathcal O_k\) and ablates second-pass decoding: (i--ii) \emph{w/o RGB/DA3 motion} removes the corresponding object cue from Eq.~\eqref{eq:object_motion_support}; and (iii) \emph{w/o release cue in merge} replaces Eq.~\eqref{eq:interval_merge} with naive length thresholding: \(M(G_k)=\mathbf 1[|G_k|\leq g_{\max}]\).
We additionally report the sensitivity test outcomes on the hyperparameters in Fig.~\ref{fig:temporal_sensitivity}, including DA3 scale $s_d$ for motion support (Eq.\eqref{eq:object_motion_support}) and thresholds from the contact and release boundary rules (Eq.\eqref{eq:second_pass_decoding}).







\begin{figure*}
    \centering
    \includegraphics[width=\linewidth]
    {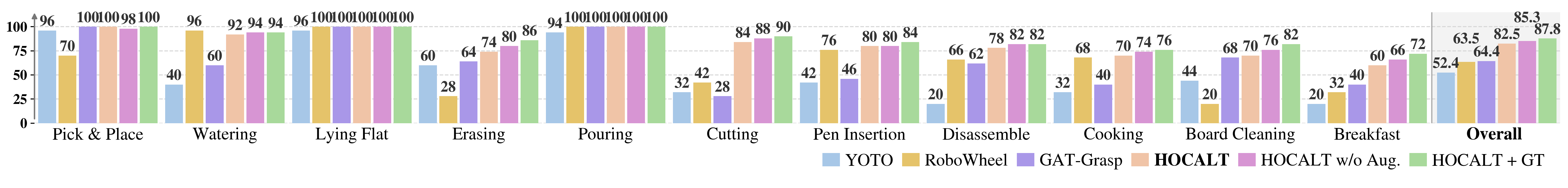}
    \vspace{-7mm}
    \caption{Per-task real-robot replay success rates (\%)}
    \label{fig:replay_success}
\end{figure*}

\noindent\textbf{Ablation results} Group A shows that removing the object-conditioned second pass produces the largest IoU decrease, from 0.85 to 0.76, indicating that the interval-specific object cue is the primary source of improvement over the hand-only first pass. 
Furthermore, Group B shows that RGB and DA3 motion provide smaller but consistent gains. In contrast, release-conditioned merging primarily improves boundary timing, reducing MAE from 11.89 to 8.34 frames and increasing SR(10) from 0.71 to 0.77. Overall, the interval-specific object cue drives most of the improvement in accuracies, while the geometry-driven components refine proposal consistency.
The sensitivity analysis in Fig.~\ref{fig:temporal_sensitivity} shows that the entry threshold $\tau_c$ has the strongest influence on localization performance. In contrast, the second-pass decoder is broadly robust to varying DA3 scale $s_d$, continuation threshold $\tau_r$, or object motion threshold $\tau_\delta$ over the tested ranges.

\begin{table}[t]
\centering

\setlength{\tabcolsep}{2pt}
\caption{Temporal localization results on EgoPAT3D and DeskTIL.}
\label{tab:egocentric_til_local}
\begin{tabular}{l|cccccccc}
\toprule
Approach & SR(3)$\uparrow$ & SR(5)$\uparrow$ & SR(10)$\uparrow$ & MAE$\downarrow$ & Rec.$\uparrow$ & IoU$\uparrow$ & Prec.$\uparrow$ & F1$\uparrow$ \\
\midrule
\multicolumn{9}{c}{\textit{EgoPAT3D (66 samples)}} \\
\midrule
\textit{EgoLoc\cite{EgoLoc}} & 0.70 & 0.87 & 0.94 & \textbf{2.45} & 0.87 & 0.78 & 0.85 & 0.85 \\
\textit{T-PIVOT\cite{wake2025open}} & 0.64 & 0.86 & \textbf{0.98} & 3.15 & \textbf{0.89} & 0.79 & 0.89 & \textbf{0.88} \\
\textit{Ours (Mono)} & \textbf{0.73} & \textbf{0.91} & 0.97 & 2.85 & \textbf{0.89} & \textbf{0.81} & \textbf{0.90} & \textbf{0.88} \\
\midrule
\multicolumn{9}{c}{\textit{DeskTIL (129 samples)}} \\
\midrule
\textit{EgoLoc\cite{EgoLoc}} & 0.72 & 0.76 & 0.93 & 3.16 & 0.94 & 0.66 & 0.67 & 0.77 \\
\textit{T-PIVOT\cite{wake2025open}} & 0.87 & 0.91 & 0.98 & 2.26 & \textbf{0.98} & 0.74 & 0.75 & 0.84 \\
\textit{Ours (Mono)} & \textbf{0.97} & \textbf{0.99} & \textbf{1.00} & \textbf{0.83} & 0.91 & \textbf{0.86} & \textbf{0.94} & \textbf{0.92} \\
\bottomrule
\end{tabular}
\end{table}

\noindent\textbf{Egocentric benchmarks}
To assess performance beyond our controlled stereo setup, we compare all methods on the public EgoPAT3D (66 videos) and DeskTIL (129 videos) samples with valid manual annotations released by EgoLoc~\cite{EgoLoc}.\footnote{\url{https://github.com/IRMVLab/EgoLoc}} The results are reported in Table~\ref{tab:egocentric_til_local}.

Across both datasets, \emph{Ours (Mono)} achieves the highest IoU and precision, while matching the best F1 on EgoPAT3D and obtaining the best F1 on DeskTIL. Overall, it shows that \emph{HOCALT} remains competitive on the monocular egocentric samples compared to VLM-based baselines.

\subsection{Real-Robot Replay Evaluation}

To remain compatible with standard visuomotor policies \cite{zhao2023learning, chi2025diffusion, ze20243d} operating on robot-centric observations, similar to \cite{YOTO}, our transferred trajectories are autonomously replayed to acquire embodiment-specific observations. 
Therefore, we aim to evaluate the transferred demonstrations by analyzing the autonomous replay success. 
A replay succeeds only if the task-specific goal is completed without intervention, and multi-stage tasks require every stage in the prescribed order. Every trajectory is evaluated five times.

\noindent\textbf{Baselines} We compare \emph{HOCALT} against following baselines on human-robot retargeting: 
(i) \emph{GAT-Grasp}~\cite{GAT-Grasp} constructs a gripper frame using the thumb--index midpoint as the grasp center; 
(ii) \emph{YOTO}~\cite{YOTO} sets the gripper center to the average position under the thumb, index, middle, and ring fingertips. Importantly, the robot motion may rely on multiple manually annotated keyframes within each interval.
For instance, \emph{Board Erasing} requires the robot to erase several board locations in one contact interval, where the keyframes are recommended to be annotated manually~\cite{YOTO}. 
(iii) \emph{RoboWheel}~\cite{zhang2026robowheel} switches between whole-hand and finger-only mapping between hand and gripper following the gesture category.
(iv) We include \emph{HOCALT + GT}, where ground-truth annotations of temporal intervals are used, and (v) \emph{HOCALT w/o Aug.}, where no LTE augmentation is applied. For \emph{HOCALT} with LTE augmentation, we uniformly sample one variant per demonstration to evaluate.

\noindent\textbf{Results} Fig.~\ref{fig:replay_success} reports an average replay success rate of $82.5\%$ for \emph{HOCALT}, substantially higher than the  \emph{RoboWheel} and \emph{GAT-Grasp} with success rates of over 60\%. \emph{YOTO} obtains lower success, particularly on tasks requiring constrained intermediate motions, where sparse keyframe interpolation can lead to execution deviations or collisions. 
Notably, \emph{HOCALT w/o Aug.} achieves $2.8$ percentage points above the augmented variant, which shows that trajectory augmentation increases the number and coverage of robot trajectories obtainable from each demonstration, while introducing a modest trade-off on the replay success.

Failure inspection reveals four failure modes in the end-to-end \emph{HOCALT} pipeline: (i) grasp retargeting failures from \emph{HOGraspFlow}; (ii) temporal localization errors that truncate or extend the transferred interaction intervals; (iii) execution failures, including object drops, unsafe contacts, and collisions; and (iv) inter-interval motion planning failures.

\subsection{Imitation Learning Policy Evaluation}
\label{sec:imitation}
\begin{table}[t]
\centering
\caption{Policy successes over 20 physical trials per task}
\resizebox{\linewidth}{!}{%
\begin{tabular}{lcccccc}
\toprule
Task
& \multicolumn{2}{c}{DP~\cite{chi2025diffusion}}
& \multicolumn{2}{c}{DP3~\cite{ze20243d}}
& \multicolumn{2}{c}{ACT~\cite{zhao2023learning}} \\
\cmidrule(lr){2-3}
\cmidrule(lr){4-5}
\cmidrule(lr){6-7}
& \emph{Ours} & \emph{Teleop}
& \emph{Ours} & \emph{Teleop}
& \emph{Ours} & \emph{Teleop} \\
\midrule
Pick \& Place
& 5/20 & 10/20 & 4/20 & 14/20 & 8/20 & 15/20 \\
Watering
& 12/20 & 14/20 & 13/20 & 10/20 & 11/20 & 16/20 \\
Lying flat
& 17/20 & 16/20 & 17/20 & 17/20 & 18/20 & 17/20 \\
Erasing
& 16/20 & 15/20 & 17/20 & 12/20 & 15/20 & 10/20 \\
Pouring
& 18/20 & 16/20 & 18/20 & 17/20 & 15/20 & 19/20 \\
Cutting
& 15/20 & 10/20 & 16/20 & 14/20 & 14/20 & 13/20 \\
Pen Insertion
& 12/20 & 13/20 & 11/20 & 13/20 & 9/20 & 10/20 \\
Disassemble
& 6/20 & 8/20 & 4/20 & 10/20 & 12/20 & 10/20 \\
Cooking
& 6/20 & 3/20 & 5/20 & 2/20 & 4/20 & 3/20 \\
Board Cleaning
& 10/20 & 8/20 & 15/20 & 11/20 & 17/20 & 13/20 \\
Breakfast
& 8/20 & 7/20 & 6/20 & 5/20 & 12/20 & 5/20 \\
\midrule
\textbf{Overall}
& \textbf{125/220} & 120/220
& \textbf{126/220} & 125/220
& \textbf{135/220} & 131/220 \\
\bottomrule
\end{tabular}%
}
\label{tab:policy_success_rate}
\end{table}

To assess whether the replayed robot demonstrations provide effective supervision relative to teleoperation, we evaluate three imitation learning algorithms: Diffusion Policy (DP)~\cite{chi2025diffusion}, 3D Diffusion Policy (DP3)~\cite{ze20243d}, and Action Chunking with Transformers (ACT)~\cite{zhao2023learning}. For each task, we construct a \emph{HOCALT} training set by uniformly sampling 50 trajectories from all successful autonomous replays. We additionally collect an equally sized teleoperation dataset using the setup shown in Fig.~\ref{fig:hardware} (D). Each policy is trained separately on both data sources for 3,000 epochs using a batch size of 128 on a single NVIDIA RTX 4090 GPU.

\noindent\textbf{Results} Table~\ref{tab:policy_success_rate} shows that policies trained on \emph{HOCALT} and teleoperated demonstrations achieve comparable overall performance across all three policy architectures. 
Despite the variations in individual tasks, the consistent aggregate performance across different architectures suggests that the replayed trajectories preserve task-relevant supervision and can be used for direct supervision of imitation learning policies. 
Thus, \emph{HOCALT} provides a scalable alternative to teleoperation, maintaining similar downstream policy performance while reducing the manual robot-control effort required for data collection.

\subsection{Human Preference Study}
\noindent\textbf{Experiment setup} To further evaluate the quality of demonstrations from \emph{HOCALT}, we conducted a blinded pairwise preference study comparing demonstrations from \textit{HOCALT} with those via teleoperation (\emph{Teleop}) across all 11 tasks.\footnote{The study only collected anonymized preferences from participants. No personally identifiable information was collected or used.}

\noindent\textbf{Protocol} For each task, we used 15 clips of successful robot executions from each method and generated 3 randomized one-to-one pairing rounds. Each clip appeared once in each round, and no clip pair was repeated across rounds. 
The resulting comparison pairs were assigned to 10 participants using a balanced assignment scheme. Each clip was evaluated by at least six distinct participants, while no participant viewed the same clip more than once. 

Before the study began, participants were instructed to base their preferences on three criteria as guidance: (i) the reasonableness of robot behavior; (ii) the stability, safety, and repeatability of the trajectory; (iii) its suitability as high-quality data for intelligent robotic agents. During each trial, the two paired clips were displayed side by side in random order. Participants were not told which method produced which clip. They indicated their preference on a scale ranging from $-100$ to $100$, where $-100$ represented a strong preference for the clip on the left, $100$ for the right one, and $0$ indicated no preference. 



\begin{figure}
\centering
    \includegraphics[width=.95\linewidth]{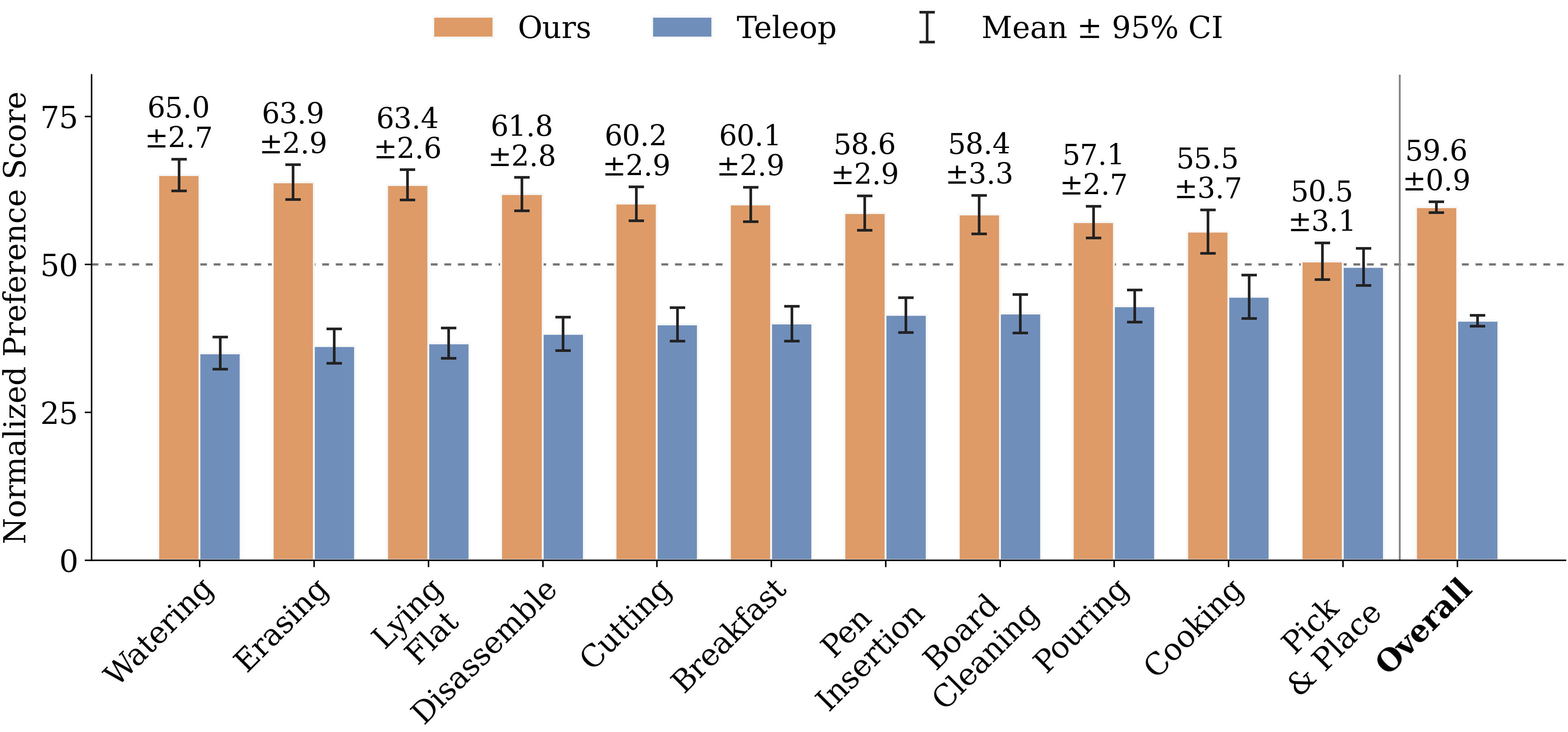}
    \vspace{-1.5mm}
    \caption{Statistics on the normalised preference between demonstrations collected from HOCALT (\emph{Ours}) and teleoperation (\emph{Teleop})}
    \label{fig:task_preference}
\end{figure}

\noindent\textbf{Results}
The results of the pairwise preference study are shown in Fig.~\ref{fig:task_preference}. Overall, participants preferred demonstrations collected using \emph{HOCALT} over those collected via \emph{Teleop}, which received an overall normalized preference score of $59.6$, with a win rate of $69.68\%$ and a non-tie win rate of $80.4\%$ 
. For the participants and clips evaluated, these results suggest that \emph{HOCALT} demonstrations were generally perceived as more reasonable and stable.

\section{Conclusions}

We presented \textit{HOCALT}, a hand-centric trajectory transfer framework that converts low-cost stereo human demonstrations into contact-aware, semantically aligned, and executable PJ end-effector trajectories. By recovering temporally consistent 3D hand motion, localizing task-relevant contact intervals, and retargeting human grasp into multi-modal PJ grasp hypotheses, \textit{HOCALT} preserves key HOI patterns during cross-embodiment transfer. Experiments demonstrate accurate temporal contact localization, reliable real-robot trajectory replay, and favorable human preference compared with demonstrations from teleoperation.

\noindent\textbf{Discussion}
The following limitations are expected to be addressed in future work: (i) We currently focus on PJ manipulation with a single active hand, although this provides a natural basis for future extension to bimanual interactions with other embodiments; (ii) Although evaluated on egocentric benchmarks, a modest extension to egocentric videos requires metric camera motion compensation, such as Structure from Motion (SfM). (iii) During trajectory transfer, we assume a rigid wrist–object transform and therefore do not model in-hand manipulation or contact forces.


\bibliographystyle{IEEEtran}
\bibliography{example}  

\end{document}